\documentclass[lettersize,journal]{IEEEtran}
\usepackage{amsmath,amsfonts}
\usepackage{algorithmic}
\usepackage{algorithm}
\usepackage{array}
\usepackage[caption=false,font=normalsize,labelfont=sf,textfont=sf]{subfig}
\usepackage{textcomp}
\usepackage{stfloats}
\usepackage{url}
\usepackage{verbatim}
\usepackage{graphicx}
\usepackage[inkscapelatex=false]{svg}
\usepackage{cite}
\usepackage{multirow}
\usepackage{booktabs}
\usepackage{bbding}
\usepackage{makecell}

\usepackage{algorithm}
\usepackage{algorithmic}
 
\makeatletter
\newcommand{\removelatexerror}{\let\@latex@error\@gobble}
\makeatother

\begin{document}

\title{Multi-scale Semantic Correlation Mining for Visible-Infrared Person Re-Identification}

\author{Ke Cheng, Xuecheng Hua, Hu Lu, Juanjuan Tu, Yuanquan Wang, Shitong Wang 
        % <-this % stops a space
\thanks{(Corresponding authors: Hu Lu and Shitong Wang.)}
\thanks{Ke Cheng, Xuecheng Hua and Juanjuan Tu are with the School of Computer, Jiangsu University of Science and Technology, Zhenjiang 212100, China (e-mail:chengke1972@just.edu.cn; huaxuecheng8888@gmail.com; ecsitu@126.com).}% <-this % stops a space
\thanks{Hu Lu is with the School of Computer Science and Communication Engineering, Jiangsu University, Zhenjiang 212013, China (e-mail:luhu@ujs.edu.cn).}
\thanks{Yuanquan Wang is with the School of Artificial Intelligence, Hebei University of Technology, Tianjin 300401, China (e-mail:wangyuanquan@scse.hebut.edu.cn).}
\thanks{Shitong Wang is with the School of Digital Media, Jiangnan University, Wuxi 214122, China (e-mail:wxwangst@aliyun.com).}
}

\markboth{Journal of \LaTeX\ Class Files,~Vol.~14, No.~8, August~2021}%
{Shell \MakeLowercase{\textit{et al.}}: A Sample Article Using IEEEtran.cls for IEEE Journals}


\maketitle

\begin{abstract}
The main challenge in the Visible-Infrared Person Re-Identification (VI-ReID) task lies in how to extract discriminative features from different modalities for matching purposes. While the existing well works primarily focus on minimizing the modal discrepancies, the modality information can not thoroughly be leveraged. To solve this problem, a Multi-scale Semantic Correlation Mining network (MSCMNet) is proposed to comprehensively exploit semantic features at multiple scales and simultaneously reduce modality information loss as small as possible in feature extraction. The proposed network contains three novel components. Firstly, after taking into account the effective utilization of modality information, the Multi-scale Information Correlation Mining Block (MIMB) is designed to explore semantic correlations across multiple scales. Secondly, in order to enrich the semantic information that MIMB can utilize, a quadruple-stream feature extractor (QFE) with non-shared parameters is specifically designed to extract information from different dimensions of the dataset. Finally, the Quadruple Center Triplet Loss (QCT) is further proposed to address the information discrepancy in the comprehensive features. Extensive experiments on the SYSU-MM01, RegDB, and LLCM datasets demonstrate that the proposed MSCMNet achieves the greatest accuracy. We have released the source code on https://github.com/Hua-XC/MSCMNet. 
\end{abstract}


\section{Introduction}
\IEEEPARstart{P}{erson} re-identification (ReID) \cite{r1wang2018transferable,r3sun2019perceive} is a technology aimed at retrieving and identifying individuals in person images captured by non-overlapping cameras. With the increasing emphasis on public safety, all-weather surveillance and retrieval systems have garnered significant attention in the field of computer vision. Existing ReID methods primarily focus on retrieving information between RGB images\cite{r6wang2018person,r8zheng2019pyramidal}. However, apart from bright daylight scenarios, the majority of online surveillance occasions should be during the night and under low lighting conditions. Therefore, the models designed for visible modality are inadequate for conducting image retrieval in low-light conditions. To address this issue, many all-weather surveillance systems incorporate infrared cameras to capture scenes in low-light environments. Nevertheless, the longer wavelength and increased scattering of infrared light result in the loss of color, texture, and detail information that is commonly present in visible images. This results in substantial discrepancies in the cross-modal. Additionally, in the same modality, there are variations in camera viewpoints, pedestrian clothing, and occlusion scenarios. \cite{r26wu2017rgb}. All of these factors present significant challenges for VI-ReID.

\begin{figure}[!t]
\centering
\includegraphics[width=3.4in]{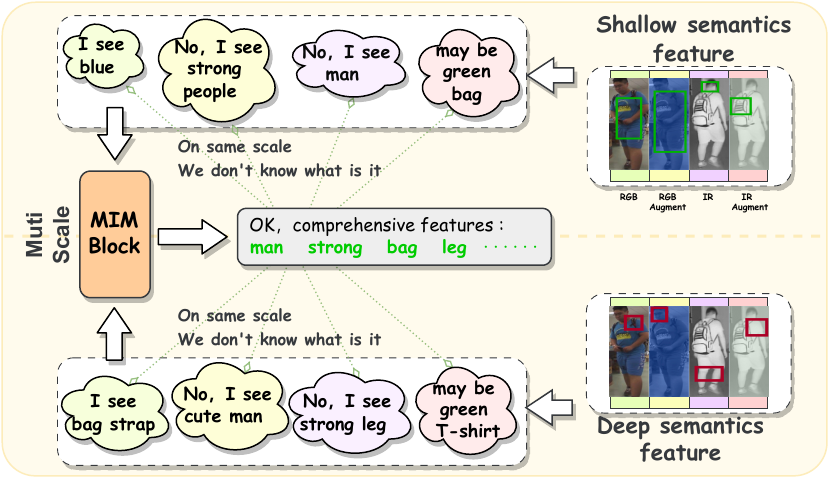}
\caption{Motivation: Due to the semantic misalignment between features within the same layer of the network, some valuable modality information cannot be utilized. However, the correlation of this semantic information can be explored at multiple scales, which enables the extraction of more comprehensive personal features. Therefore, conducting a multi-scale exploration of valuable semantic information is crucial. Our MSCMNet effectively achieves this objective.}
\label{Fig_1}
\end{figure}

Currently, the main challenges faced by the existing models can be categorized into three aspects:  1) The infrared modality lacks color, texture, and other detailed information, making it difficult to fully extract modal-shared features from the visible modality, which contains more comprehensive information. 2) During the process of feature extraction, modality information loss occurs, making it challenging for the network to make full use of the information within the dataset. 3) Significant variations in poses, attire, and backgrounds for the same identity under different cameras further hinder the extraction of effective and robust features.

Existing outstanding works have been proposed in VI-ReID, mainly concentrated on utilizing a dual-stream network to extract the cross-modal shared features. Specifically, there are many works using data augmentation for RGB images or generating auxiliary modality to eliminate the impact of effects such as color and details on the visible images. This approach enables the network to focus on characteristic features, such as body shape and person outline, minimizing the discrepancies between visible and infrared images. While these methods can effectively handle the differences between modalities, they still cannot avoid the loss of valuable information during the feature extraction process. The lost information typically represents modality-specific features. They are temporarily invaluable at the same scale due to semantic inconsistencies. However, As shown in Fig. 1, as modality information is extracted in deep networks, the possibility arises that lost modality-specific semantic information in shallow layers can be utilized in deep layers.  Therefore, the ability to take advantage of all information contained within cross-modality has become more and more important for evaluating the performance of the model.

Furthermore, in comparison to existing multi-scale approaches, our work innovatively considers the deep exploration of modality information and designs a novel structure that utilizes multiple-scale features, which enables the extraction of more comprehensive features.

Concretely, we propose a novel deep learning framework named Multi-scale Semantic Correlation Mining Network (MSCMNet). This network is composed of Quadruple-Stream Feature Extractors (QFE) and Multi-scale Information Correlation Mining Block (MIMB). The innovative QFE captures modality features from channel augmented dimensions and image global dimensions in the way of non-shared parameters. In addition, it performs feature fusion during testing. We apply channel augmentation techniques to expand the dimensions of RGB\cite{r38ye2021channel} and IR images. Then, the proposed Multi-scale Information Correlation Mining Block (MIMB) effectively addresses the issue of valuable modality information missing during feature extraction. Moreover, it mines the potential correlations between different stream features at multiple scales and explores implicit semantic correlation among cross-modal features. Finally, we design the Quadruple-Center Triplet Loss (QCT) as a novel learning objective that enhances the comprehensiveness of modality-shared features within multidimensional feature spaces.

The contributions of our work are summarized as follows:
\begin{itemize}
\item{A novel deep learning framework (MSCMNet) is proposed to focus on extracting comprehensive modality features.}
\item{There are two modules in the proposed MSCMNet. 1) The QFE is enabled to extract broader semantic information. 2) The MIMB explores the correlations of semantic information across muti-scales and effectively reduces information loss during the feature extraction process.}
\item{A novel Quadruple Center Triplet Loss is coined to handle the semantic information discrepancy within cross-modal features.}
\item{Our proposed MSCMNet outperforms other state-of-the-art methods in the VI-RelD task, as demonstrated by extensive evaluations on the SYSU-MM01, RegDB, and LLCM datasets.}
\end{itemize}

The remainder of this paper is organized as follows: some works related to person re-identification and visible infrared person ReID are introduced briefly in Section II, and the details of the proposed MSCMNet are presented in Section III. Following that, experiment settings and experimental results are reported in Section IV. Finally, the conclusion is drawn in Section V.

\section{Related Work}
\subsection{Person Re-identification}
Person re-identification aims to match person images captured by different cameras during daylight. With the emergence of deep convolutional neural networks (CNNs), there has been a great effort in this field to design effective loss functions \cite{r4zhou2019discriminative,r5chen2019abd,r6wang2018person,r7xia2019second,r8zheng2019pyramidal}. These loss functions are crafted to constrain features and enforce relationships between individuals of distinct identities, employing various strategies such as triplet constraints \cite{r13wang2016dari}, quadruplet constraints \cite{r14chen2017beyond}, and group consistency constraints \cite{r15chen2018group}. These methods, which learn features from entire person images, often encounter challenges from intra-modal variations, specifically those induced by changes in human body posture and partial occlusion \cite{r17varior2016siamese}. As a result, alternative approaches have arisen to address these issues. Some of these methods focus on fine-grained person image descriptions, achieved by either partitioning body parts uniformly \cite{r16fu2019horizontal,r17varior2016siamese,r18wang2018learning,r19li2018harmonious,r8zheng2019pyramidal} or employing attention mechanisms \cite{r20liu2017end,r21liu2017hydraplus,r23zhao2017deeply,r12zheng2019re}. Moreover, certain strategies incorporate both local and global information to enhance performance \cite{r1wang2018transferable,r2sun2018beyond,r3sun2019perceive}. However, the practical reality is that most cameras transition between visible and infrared modalities during both day and night. Owing to the significant cross-modal disparities, the efficacy of single-modal solutions is severely compromised in the context of VI-ReID tasks, leading to suboptimal generalization performance.
\subsection{ Visible-Infrared Person ReID}
In addressing the challenge of cross-modal discrepancy, Zero-Padding \cite{r26wu2017rgb} embarked on a pioneering effort by introducing a zero-padding single-stream network and contributing to the SYSU-MM01 dataset. Dual-Stream network\cite{r27ye2019bi} designed to match facial images under two different models.
Some methods based on metric learning \cite{r32liu2020parameter,r31zhu2020hetero}, which introduced the Heterogeneous Center (HC) loss to mitigate modality gaps\cite{r31zhu2020hetero} and  \cite{r32liu2020parameter} further devised the Heterogeneous Center Triplet loss. At the same time, \cite{r30zhang2021global,r34sun2022notDCL,r35park2021learningLbA} Combining global and local information for fine-grained learning. Furthermore, the method based on GAN and data augmentation is dedicated to reducing the discrepancy between cross-modality\cite{r43dai2018cross,r44wang2020cross,r45AliGANwang2019rgb}. \cite{r46li2020infrared,r37zhao2021joint,r38ye2021channel} based on auxiliary modalities and data augmentation. \cite{r46li2020infrared}introducing an auxiliary X-modality. \cite{r37zhao2021joint} eliminating the influence of RGB color information. Effective data augmentation techniques \cite{r38ye2021channel} and some Multi-scale exploration work was proposed\cite{zhang2021multiMMM1,liu2022learnMMM2,wen2022crossMM3}, but those works inadequately considered the disparities in channel augmentation and images during the process of feature extraction. One of the motivations for our work is to take full advantage of the semantic information discrepancy brought by different data augmentation in the same modality. In addition, there are some advanced works. The transformer framework was employed in\cite{rPMTlu2023learning,he2021transreid,zhao2022spatial,liang2023cross}. DEEN\cite{r47zhang2023diverse} generates diverse features to learn rich information feature representations for bridging cross-modal disparities. However, the feature generation of DEEN via convolution makes it difficult to control the information retained by the features. \cite{feng2022occluded} is a matching framework for occlusion scenes. PartMix\cite{r52kim2023partmix} is proposed to enhance the sample. Finally, works related to semantic information. \cite{r50wu2021discover} exploited nuanced but discriminative information by a proposed pattern alignment module and a modality alleviation module. DSCNet\cite{r51zhang2022dual} optimizes channel consistency from two aspects, fine-grained inter-channel semantics and comprehensive intermodality semantics. \cite{r48feng2023shape}Ensure learning eliminates semantic concepts related to body shape from the features. 

\begin{figure*}[!t]
\centering
\includegraphics[width=7in]{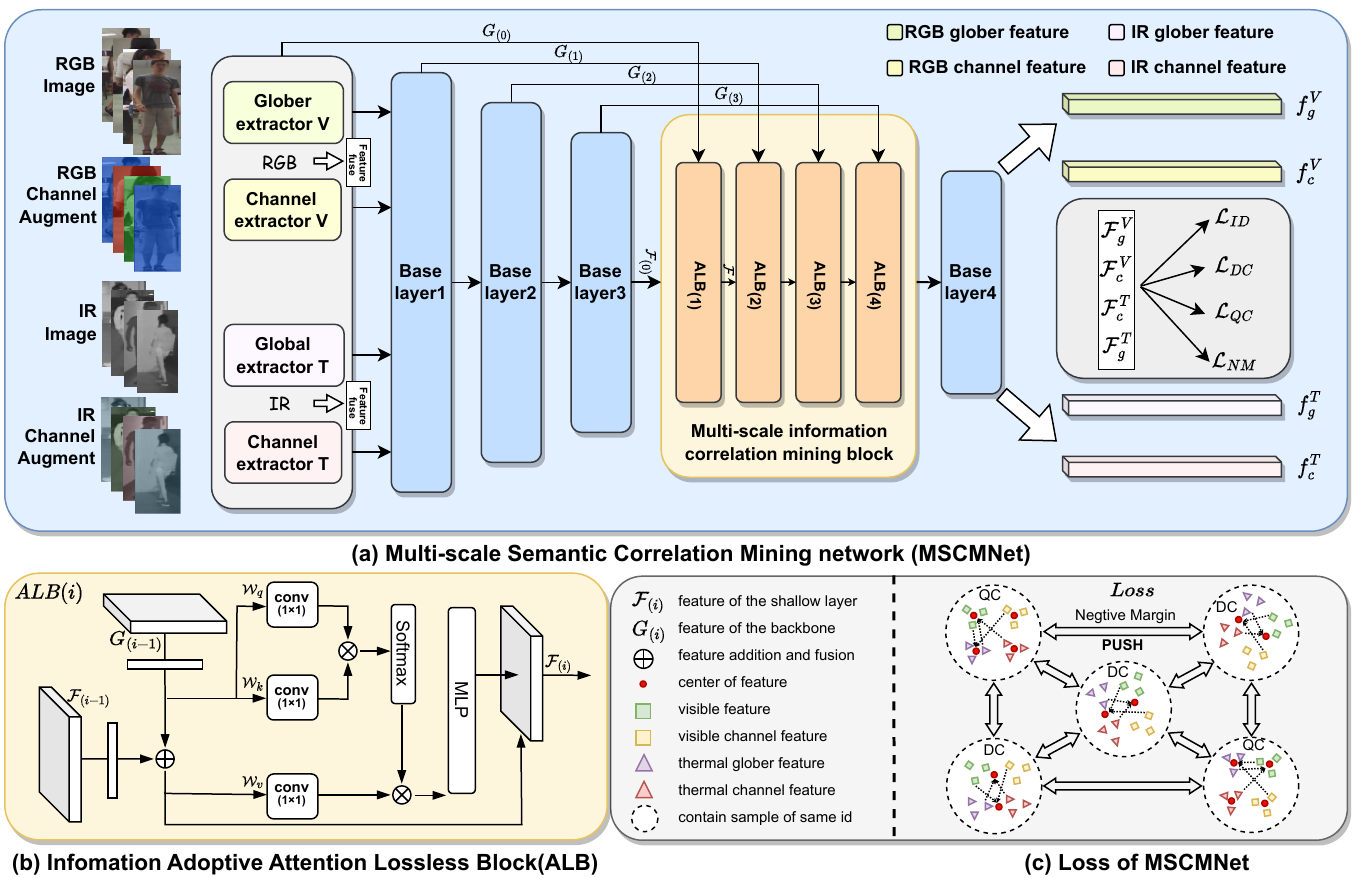}%
\hfil
\caption{The framework of our proposed method: (a) The overall structure of visible-infrared Multi-scale Semantic Correlation Mining network (MSCMNet), which is based on the Residual Neural Network. MSCMNet contains three components: Quadruple-Stream Feature Extractors (QFE), Multi-scale Information Correlation Mining Block (MIMB), and the total loss. (b) The conceptual illustration of the designed Information Adoptive Lossless Block (ALB), which is the component of MIMB. (c) The diagram of the loss function contains quadruple center loss (QC), dual center loss (DC), and negative margin loss (NM).}
\label{fig_sim}
\end{figure*}

While the above methods have proven effective. However, they have not adequately taken into account the varying semantic information emphasized by the global image and the image channel. This deficiency has led to incomplete extraction of modality features. Consequently, we propose a quadruple-stream network that performs image feature extraction at both the channel and image global levels. After that, we facilitate the fusion of semantic information across different dimensions within the features. This approach enables the exploration of implicit correlations among cross-modal features and the extraction of comprehensive modal features.

\section{The Proposed Method}
In this section, we provide a comprehensive exposition of the Multi-scale Semantic Correlation Mining Network (MSCMNet) proposed for VI-ReID, as illustrated in Fig. 2(a). Initially, we apply channel-level data augmentation to the dataset, compelling the Quadruple-Stream Feature Extractors to capture diverse modality-specific semantic features. This extractor is an innovative approach for extracting modality-specific features. Subsequently, the Multi-Scale Information Correlation Mining Block is dedicated to further exploring the implicit correlation of semantic information across multiple scales and reducing information loss during the feature extraction process. In contrast to existing multi-scale exploration methods, we have devised a novel multi-scale architecture. Finally, the Quadruple Center Triplet Loss is to handle the information discrepancy present in the quadruple-stream features. The following sections provide detailed explanations of our Quadruple-Stream Feature Extractor (QFE, \S\ \uppercase\expandafter{\romannumeral2}-B), Multi-Scale Information Correlation Mining Block (MIMB, \S\ \uppercase\expandafter{\romannumeral2}-C), and Quadruple Center Triplet Loss (QCTloss, \S\ \uppercase\expandafter{\romannumeral2}-D).

\subsection{Problem Formulation}
The RGB global images exhibit details such as color, texture, and other fine-grained features. In contrast, IR images are single-channel and inherently lack information such as color and texture.  However, for each channel of the RGB modality, the emphasis is on expressing the inherent features of individuals, such as body posture and outline, rather than focusing on color or details. In order to achieve semantic alignment between IR images and RGB images, we enhance the IR modality by giving some channel information. This enables the IR images to be similar to RGB images by having three channels, and each channel contains distinct information. Regarding RGB images, we employ channel exchange \cite{r38ye2021channel} to select channel semantic information. As a result, we acquire datasets that emphasize distinct semantic information. Formally, the visible and the infrared images can be formulated as $[\mathcal{X}^\mathcal{V},\mathcal{X}^\mathcal{T}]$
We expand the images in the RGB dataset through data augmentation to create a set of global level images $\{\mathcal{X}_{id}^{\mathcal{V}_{g}}\}^N_{i=1}$ and channel level images$\{\mathcal{X}_{id}^{\mathcal{V}_{c}}\}^N_{i=1}$. Similarly, the IR global level images $\{\mathcal{X}_{id}^{\mathcal{T}_{g}}\}^N_{i=1}$  and channel level images $\{\mathcal{X}_{id}^{\mathcal{T}_{c}}\}^N_{i=1}$. we set them as $[\mathcal{X}_i^{\mathcal{V}_g},\mathcal{X}_i^{\mathcal{V}_c},\mathcal{X}_i^{\mathcal{T}_g},\mathcal{X}_i^{\mathcal{T}_c}]$. The dimensions of $\mathcal{X}_i\in \mathbb{R}^{\left[B,C,H,W\right]}$. 

In this context, $\mathcal{X}$ represents the dataset, while $i$ denotes the identity of pedestrians. Then $\mathcal{V}$ and $\mathcal{T}$ respectively stand for the visible modality and the infrared modality. $g$ and $c$ refer to the global-level image and the channel-level image. $N$ signifies the total number of identities. $B$ represents the batch size, $C$ is the number of channels in the image, and $H$ and $W$ denote the height and width of the image. The abundance of semantic information facilitates the network in more effectively uncovering the relevance of semantics.

\begin{figure}[!t]
\centering
\includegraphics[width=3.4in]{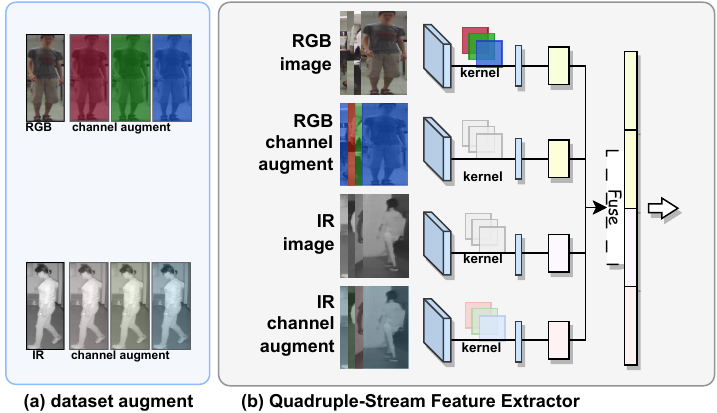}
\caption{(a) illustrates the extended dataset through data augmentation. (b) illustrates our quadruple-stream feature extractor, consisting of four convolutional layers with non-shared parameters.}
\label{Fig_3}
\end{figure}

\subsection{Quadruple Feature Extraction and Fusion}
\subsubsection{Quadruple-Stream Feature Extractor}  
As shown in Fig. 3(b), the quadruple-stream feature extractor is designed to separate the features emphasizing color and detail from those emphasizing the shape and outline of the person. In the previous dual-stream networks that utilized RGB data augmentation, these works just extracted shared features across different semantic dimensions, leading to the loss of specific semantic information generated by data augmentation. Therefore, we hypothesize that certain features in the shallow layers of the network may temporarily exhibit semantic inconsistencies due to network depth, making them unrecognizable as shared features. However, as the network deepens, the extracted features become increasingly refined. Between the useless semantic information from the shallow layers and the more comprehensive features in the deep layers, we may explore new potentially shared features. Compared to the existing dual-stream network, we propose the quadruple-stream feature extractor, which extracts features containing diverse semantic dimension information in both augmented and original images without sharing parameters.

Specifically, given the input of the augmented dataset 
$[\mathcal{X}_i^{\mathcal{V}_g},\mathcal{X}_i^{\mathcal{V}_c},\mathcal{X}_i^{\mathcal{T}_g},\mathcal{X}_i^{\mathcal{T}_c}]$. For each branch of the input, we perform feature extraction using the Quadruple-Stream Feature Extractor $\mathcal{E}\left(\cdot\right)$.

\begin{equation}
\label{1}
[G_{i}^{\mathcal{V}_g},G_{i}^{\mathcal{V}_c},G_{i}^{\mathcal{T}_g},G_{i}^{\mathcal{T}_c}]
=\mathcal{E}([{\mathcal{X}}_i^{\mathcal{V}_g},
{\mathcal{X}}_i^{\mathcal{V}_c},
{\mathcal{X}}_i^{\mathcal{T}_g},
{\mathcal{X}}_i^{\mathcal{T}_c}]
)
\end{equation}

\begin{equation}
\label{2}
G_{(0)}= [G_{i}^{\mathcal{V}_g},G_{i}^{\mathcal{V}_c},G_{i}^{\mathcal{T}_g},G_{i}^{\mathcal{T}_c}]
\end{equation}
$G$ represents the features extracted by the network, and the subscript $\left( 0 \right)$ denotes that the features are extracted by the $0\text{-}th$  layer of the network, which refers to our quadruple-stream feature extractors. The output dimension of quadruple-steam feature extractor $G\in \mathrm{R}^{2*B\times C\times H\times W}$

\subsubsection{Quadruple Feature Fusion}  
During the testing phase, we integrate features from different branches within the same modality. However, during the training process, we do not perform feature fusion. This decision is based on our intention to encourage the network to extract more diverse modality-specific features using four separate feature extractors rather than focusing on shared features. In this fusion module, we merge the features extracted by the four separate feature extractors into two categories: visible and infrared.

 Specifically, after obtaining the features extracted by the extractor $[G_{i}^{\mathcal{V}_g},G_{i}^{\mathcal{V}_c},G_{i}^{\mathcal{T}_g},G_{i}^{\mathcal{T}_c}]$, we perform feature fusion on these features.

\begin{equation}
\label{3}
\begin{aligned}
&G_i^{\mathcal{V}_m}={\alpha \cdot G}_i^{\mathcal{V}_g}\oplus{\left(1-\alpha\right)\cdot G}_i^{\mathcal{V}_c}\\
&G_i^{\mathcal{T}_m}={\alpha \cdot G}_i^{\mathcal{T}_g}\oplus{\left(1-\alpha\right)\cdot G}_i^{\mathcal{T}_c}
\end{aligned}
\end{equation}
The fused features representing the fusion of the visible and infrared modalities are denoted as $G_i^{\mathcal{V}_m}$ and $G_i^{\mathcal{T}_m}$ respectively. Here, $m$ signifies that the features have been formed through fusion and $\mathcal{V}$ and $\mathcal{T}$ indicate the respective modalities. Therefore, during training, the output of the feature extractor is $[G_{i}^{\mathcal{V}_g},G_{i}^{\mathcal{V}_c},G_{i}^{\mathcal{T}_g},G_{i}^{\mathcal{T}_c}]$, while in testing mode, the output of this module is $[G_i^{\mathcal{V}_m} , G_i^{\mathcal{T}_m}]$. We define the parameters $\alpha = 0.5$.

\subsection{Multi-scale Information Correlation Mining Block}
The key to improving our model performance lies in effectively utilizing the features extracted by the four-stream feature extractors. Following the feature extraction through the shared parameter layers, we obtain shared features across different semantic dimensions. With the continuous increase in network depth, the extracted features become increasingly refined. Therefore, our objective is to integrate these refined features and features containing rich semantic information from shallow layers. The purpose of this fusion is to enrich the feature information present in the deep layers of the network. However, it is crucial to acknowledge that this fusion incorporates both semantic relevant and irrelevant information. To address this, we design ALB to guide the network mining of truly effective features.
Finally, the aim of this module is to possess the following functionalities:

\begin{enumerate}
\item{Filtering out genuinely valuable information and further extracting comprehensive features.}
\item{Reusing the lost semantic information in the shallow layers of the network and exploring the implicit semantic correlations between the shallow-layer features and the deep-layer features.}
\end{enumerate}

To accomplish this, we set the features extracted from each layer of the backbone as $G_{i}$, The term $i$ denotes the feature output of the $i\text{-}th$ layer. To distinguish the output of the layer from the MIMB module, we designate the feature output obtained from the $3\text{-}th$ layer as our input to the MIMB module, denoted as $\mathcal{F}_0$.

We combine four ALB layers to form the MIMB module by designing a novel multi-scale structure. As shown in Fig. 2(b), each ALB has inputs $\mathcal{F}_i$ and $G_i$ . Here, $\mathcal{F}$ is the input of ALB in the backbone, which represents deep-layer features. While $G$ is the preserved output in the shallow network, representing the lost information. We should aim to explore the relationships between shallow-level semantics and deep-level semantic information across multiple scales. As such, we design a novel multi-scale structure\cite{r55gao2019res2net,r53lin2017feature} to address this issue. To enable the utilization of this lost information, we apply a Convolutional Layer and regularization to $\mathcal{F}$ and $G$, mapping them to the same feature space. We also fuse the dimensionally reduced features. Thus, we define the $\psi$:

\begin{equation}
\label{4}
\psi_i^Q=F_{convN}(\  G_i \  )
\end{equation}
\begin{equation}
\label{5}
\psi_i^K=F_{convN}(\  G_i \  )
\end{equation}
\begin{equation}
\label{6}
\psi_i^V=\alpha\cdot F_{convN}(G_i)\ \oplus \ (1-\alpha)\cdot F_{convN}(\mathcal{F}_i)
\end{equation}

Where $F_{convN}$ is to perform dimensionality reduction and we set $\alpha = 0.1$. In order to focus on the effective parts of this information, we adopt a multi-head attention mechanism \cite{r54vaswani2017attention} module as a basic block. This ALB module enables to further explore the valuable part of the shallow layer information and mine the semantic correlation of these features at Multi-scale. Specifically, we denote $\psi_i^Q$, $ \psi_i^K$, $\psi_i^V$ as the query, key, and value inputs to the attention mechanism, respectively.

\begin{equation}
\label{7}
\mathcal{F}_{i+1}=F_{conv}\left(\mathcal{A}\left(\psi_i^Q,\psi_i^K,\psi_i^V\right)+\mathcal{F}_i\right)
\end{equation}

Where the attention mechanism, denoted as $\mathcal{A}$, consists of multi-head attention and a single-layer MLP. Subsequently, we apply a convolutional layer, denoted as $F_{conv}$, to increase the dimensionality of the features after the attention mechanism, enabling subsequent ALB modules to utilize these features. Additionally, we incorporate residual connections into the ALB to further enhance its effectiveness.

\begin{figure}[!t]
\centering
\includegraphics[width=3.5in]{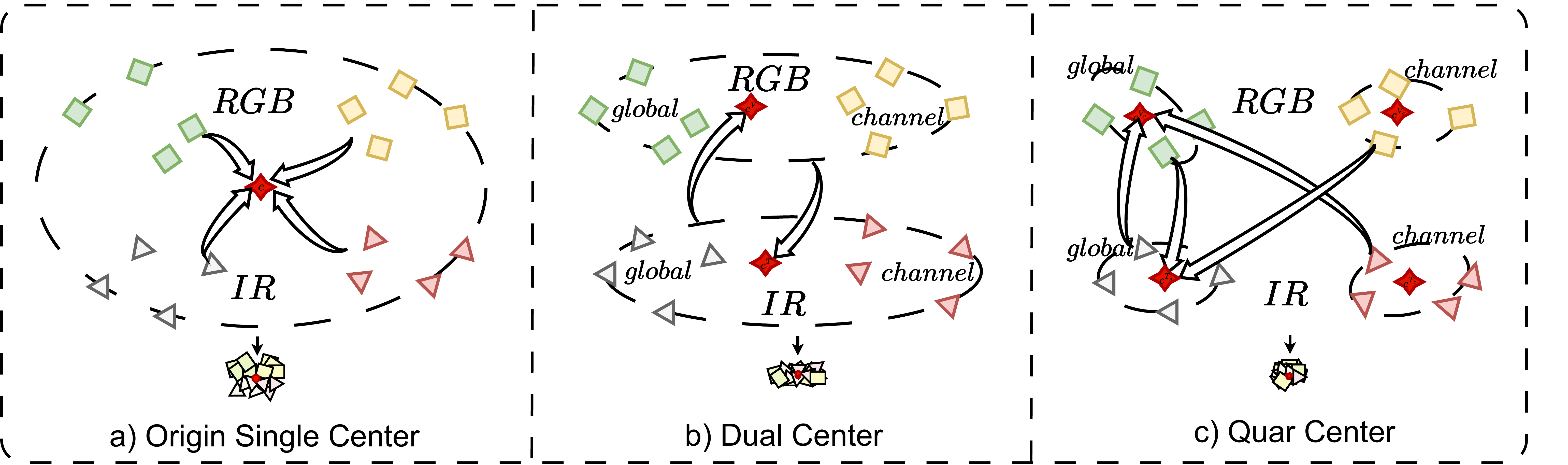}%
\hfil
\caption{The QC Loss consists of the dual center loss (b) and quar center loss (c). The dual center is employed to alleviate the significant differences between modalities, while the quar center enables the comprehensive utilization of information contained within the four-stream network.}
\label{fig_4}
\end{figure}

\subsection{Overall loss function}
In our Quadruple Center Triplet Loss, we aim to fully exploit the rich semantic information contained within the features. consider intra-modality features, the primary focus is on accurately capturing the discriminative information of persons. As for cross-modality features, we strive to explore the semantic correlations and ensure that modality-specific information can also be utilized effectively. To achieve this objective, we propose dual-center loss and quar-center loss to execute intra-modal and inter-modal constraints. Additionally, we introduce a negative boundary loss to constrain the distances between feature centers of different identities. 

As shown in Fig. 4, in contrast to origin center loss \cite{r31zhu2020hetero,r32liu2020parameter} and its existing variations that aggregate intra-modality features towards their respective centers, we directly guide the features of one modality toward the center of another modality to establish a cross-modal cross-aggregation scenario. This idea allows the features to fit directly to the center of the other modality, rather than focusing on the proximity between the centers, resulting in stronger constraint effects. It also preserves as much diverse dimensional information and modality-specific details as possible within the features.

\subsubsection{Quadruple multi-dimensional Enhancement Loss}  
From the network model, we obtain four sets of features, denoted as $f_i$. We partition these features into four separate sets, each representing a different dimensional feature.

\begin{equation}
\label{7}
\left[f_i^{\mathcal{V}_g},f_i^{\mathcal{V}_c},f_i^{\mathcal{T}_g},f_i^{\mathcal{T}_c}\right]=f_i
\end{equation}

Through computations, we derive the centers for the four-steam feature.

\begin{equation}
\label{8}
[c^{\mathcal{V}_g},c^{\mathcal{V}_c},c^{\mathcal{T}_g},c^{\mathcal{T}_c}]_i
=
\frac{1}{K}(\sum_{p=1}^{K}[
f_p^{\mathcal{V}_g},
f_p^{\mathcal{V}_c},
f_p^{\mathcal{T}_g},
f_p^{\mathcal{T}_c}]_i)
\end{equation}

We denote $c$ represents the center of the samples under different streams of the network. $p$ denotes all the samples under a particular identity. $\mathcal{V}$ and $\mathcal{T}$ refers to the visible light modality and the infrared modality. The superscript $g$ and $c$ represent the global feature and channel feature, respectively.

 We believe that global information contains more stable and reliable feature representations. Therefore, we use the center of the modality global feature as the reference point. As shown in Fig. 4(b). We encourage the modality features to approach the center of another modality. We do not utilize center constraints within the same modality. This is because the intra-modality constraints tend to prioritize the extraction of shared features among distinct semantic information. This leads to the loss of diverse semantic information within the same modality.

\begin{equation}
\begin{split}
\label{9}
\mathcal{D}^{quar}_i=&{\parallel{f_i^{\mathcal{V}_g}-c_i^{\mathcal{T}_g}} \parallel}_2+
{\parallel{f_i^{\mathcal{V}_c}-c_i^{\mathcal{T}_g}} \parallel}_2\\+
&{\parallel{f_i^{\mathcal{T}_g}-c_i^{\mathcal{V}_g}} \parallel}_2+
{\parallel{f_i^{\mathcal{T}_c}-c_i^{\mathcal{V}_g}} \parallel}_2
\end{split}
\end{equation}

In which $\mathcal{D} $ represents the Euclidean distance measure obtained by summing the squared distances between the four centers. Here, $\parallel\cdot \parallel$ denotes the $L2$ norm (Euclidean Distance).

\subsubsection{Modality-information Enhancement Loss}  
Similarly, we decompose the four sets of features, denoted as $f_i$, into the overall features of the two modalities.
\begin{equation}
\label{10}
\left[f_i^{\mathcal{V}},f_i^{\mathcal{T}}\right]=f_i
\end{equation}

By performing computations, we determine the centers of the modalities, which are composed of features from the global $g$ and channel $c$ dimensions.

\begin{equation}
\label{11}
[c^\mathcal{V},c^\mathcal{T}]_i
=\frac{1}{2K}(\sum_{p=1}^{K}[f_p^{\mathcal{V}_g}+f_p^{\mathcal{V}_c} ,f_p^{\mathcal{T}_g}+f_p^{\mathcal{T}_c}]_i)
\end{equation}

As shown in Fig. 4(c), we constrain the modality features with the center of another modality using the $L2$ norm.

\begin{equation}
\label{12}
\mathcal{D}^{dual}_i={\parallel{f_i^\mathcal{V}-c_i^\mathcal{T}} \parallel}_2+
{\parallel{f_i^\mathcal{T}-c_i^\mathcal{V}} \parallel}_2
\end{equation}

\begin{figure}[!t]
    \label{HHH}
    \removelatexerror
    \begin{algorithm}[H]
        \caption{The training procedure of MSCMNet}
        \begin{algorithmic}[1]
            \REQUIRE Training set: $\mathcal{X}_{train}$          %%input
            \ENSURE Network parameters $\Theta_1$, $\Theta_2$, $\Theta_3$ %%output
            \STATE {Obtain $[\mathcal{X}^{\mathcal{V}_g},\mathcal{X}^{\mathcal{V}_c},\mathcal{X}^{\mathcal{T}_g},\mathcal{X}^{\mathcal{T}_c}]$ from datasets $\mathcal{X}_{train}$.}  
            \STATE {Pretrain ResNet50 on ImageNet using parameters $\Theta_1$, initialize the QFE parameters $\Theta_2$, MIMB parameters $\Theta_2$.}  
            \STATE {\textbf{while} not convergence \textbf{do}}  
            \STATE \quad{Obtain $G_i$ by QFE and the $i\text{-}th$ layer of Baseline. \\ \quad// (1)–(3)}  
            \STATE \quad{Obtain $\mathcal{F}_i$ by MIMB. // (4)–(7)}  
            \STATE \quad{Obtain $[f^{\mathcal{V}_g} ,f^{\mathcal{V}_c} ,f^{\mathcal{T}_g} ,f^{\mathcal{T}_c}]$ by joint exploration of \\ \quad$\mathcal{F}_i$ and $G_i$.  }  
            \STATE \quad{Aggregate $f^{\mathcal{V}_g} ,f^{\mathcal{V}_c} ,f^{\mathcal{T}_g}$ and $ ,f^{\mathcal{T}_c}$. // (8)–(14) } 
            \STATE \quad{Calculate $\mathcal{L}_{ID}$ and $\mathcal{L}_{QCT}$ by (15),(16).} 
            \STATE \quad{Optimize parameters $\Theta_1$, $\Theta_2$, $\Theta_3$ according to (17).} 
            \STATE {\textbf{end while}}

        \end{algorithmic}
    \end{algorithm}
\end{figure}

\subsubsection{Negative Bounder margin loss}  
In the previous context, we obtain the sample centers and aggregate the features of the same ID together. However, constraining features within the same identity alone are not sufficient. We also need to enforce distance constraints between different identities to ensure distinguishability among samples from different IDs. Given that the center loss mentioned above already provides effective constraints for positive samples within the same identity, we only need to apply the Negative Bounder loss to constrain negative samples between different identities. Thus, As shown in Fig. 1(c). we set a lower bound $ \rho $ for the distances between different IDs.

\begin{equation}
\label{14}
\mathcal{L}_{NM}={{{\sum^{N}_{\mathop{i,j=1}\limits_{\forall i\neq j}}}}}{[\ \rho\ -\  \parallel f_i-c_{y_j}\parallel_2  \ ]_+}
\end{equation}

In fact, we impose the constraints on each feature $f_i$ within each identity (ID), ensuring that its distance to the feature center $c_{y_i}$ of other identities remains above a threshold of $\rho$.

\subsubsection{Objective Function}
In the previous context, the designed loss functions enable the RGB and IR modality features to incorporate more modality-specific information and explore the underlying correlations between features. The quadruple-center is employed to constrain modality-specific information, while the dual-center provides overall constraints between modalities. However, if the features contain excessive uncorrelated modality-specific information, it can hinder the fitting of our model. Conversely, if there is an insufficient amount of effective modality-specific features, our network may degrade into a dual-stream network. Therefore, we carefully balance the degrees of quadruple-center constraints and dual-center constraints to get a more comprehensive modal center.
\begin{equation}
\label{13}
\mathcal{L}_{QC}=\alpha\sum_{i=1}^{N}\mathcal{D}^{quar}_i+(1-\alpha)\sum_{i=1}^{N}\mathcal{D}^{dual}_i
\end{equation}

The parameter $\alpha$ is set to $0.05$ in order to balance the terms dual-center and quadruple-center. Finally, we also incorporate the ID loss. The overall loss function is defined as follows:
\begin{equation}
\label{15}
\mathcal{L}_{QCT} = \mathcal{L}_{QC}+\mathcal{L}_{NM}
\end{equation}
\begin{equation}
\label{16}
\mathcal{L}_{total} = \mathcal{L}_{ID}+\mathcal{L}_{QCT}
\end{equation}

 The training process of our approach is illustrated in Algorithm 1.

\section{Experimental Result}

\subsection{Datasets and Evaluation Metrics}
The SYSU-MM01 dataset \cite{r26wu2017rgb} provides 86,628 visible images and 15,792 infrared images captured by four visible cameras and two infrared cameras. It consists of 491 individual identities and includes both all-search and indoor-search modes. We use 3,803 infrared images as the query set and randomly select 301 images from other visible images as the gallery set.

The RegDB dataset \cite{RRegDBnguyen2017person} contains 4,120 paired visible and infrared images captured by overlapping cameras. It consists of 412 identities, with each identity having 10 visible images and 10 infrared images. For training, we randomly select all images from 206 identities, while the remaining 206 identities are used for testing.

The LLCM \cite{r47zhang2023diverse} dataset utilizes a 9-camera network deployed in low-light environments, which can capture the VIS images in the daytime and capture the IR images at night. This dataset contains 46,767 bounding boxes of 1,064 identities, encompassing various climate conditions and clothing styles.

We utilize Cumulative Matching Characteristics (CMC), mean Average Precision (mAP), and mean Inverse Negative Penalty (mINP)\cite{rAGWye2021deep} as our primary evaluation metrics.
\begin{table*}\footnotesize
    \centering
    \caption{Comparison With The State-Of-The-Arts On SYSU-MM01 Dataset. Rank-k Accuracy (\%) and MAP (\%) Are Reported.}
    \setlength{\tabcolsep}{3.8mm}{
    \begin{tabular}{l|c|cccc|cccc}   
        \toprule
        \multirow{2}*{Method} & \multirow{2}*{Publish} & \multicolumn{4}{c}{All Search} & \multicolumn{4}{c}{Indoor search} \\
          &       & Rank-1  & Rank-10  & Rank-20 & mAP & Rank-1  & Rank-10  & Rank-20 & mAP \\          
        \midrule
        Zero-Pad \cite{r26wu2017rgb}            & ICCV 17  & 14.80 & 54.12 & 71.33 & 15.95 & 20.58 & 68.38 & 85.79 & 26.92 \\
        HCML  \cite{rHCMLye2018hierarchical}    & AAAI 18  & 14.32 & 53.16 & 69.17 & 16.16 & 24.52 & 73.25 & 86.73 & 30.08 \\
        AliGAN \cite{r45AliGANwang2019rgb}      & ICCV 19  & 42.40 & 85.00 & 93.70 & 40.70 & 45.90 & 87.60 & 94.40 & 54.30 \\
        DDAG  \cite{r41ye2020dynamicDDAG}       & ECCV 20  & 54.75 & 90.39 & 95.81 & 53.02 & 61.02 & 94.06 & 98.41 & 67.98 \\ 
        AGW  \cite{rAGWye2021deep}              & TPAMI 21 & 47.50 & 84.39 & 92.14 & 47.65 & 54.17 & 91.14 & 95.98 & 62.97 \\
        LbA  \cite{r35park2021learningLbA}      & ICCV 21  & 55.41 &   -   &   -   & 54.14 & 58.46	&   -   &   -	& 66.33 \\
        CAJ  \cite{r38ye2021channel}            & CVPR 21  & 69.88 & 95.71 & 98.46 & 66.89 & 76.26 & 97.88 & 99.49 & 80.37\\
         \midrule
        DML    \cite{RDMLzhang2022dual}         & TCSVT 22 & 58.40 & 91.20 & 95.80 & 56.10 & 62.40 & 95.20 & 98.70 & 69.50\\
        SPOT \cite{rsportchen2022structure}     & TIP 22   & 65.34 & 92.73 & 97.04 & 62.25 & 69.42 & 96.22 & 99.12 & 74.63 \\
        FMCNet  \cite{r39zhang2022fmcnet}       & CVPR 22  & 66.34 &   -   &	-   & 62.51 & 68.15	&   -   &   -	& 74.09 \\
        DCLNet  \cite{r34sun2022notDCL}         & ACM MM 22& 70.80 &   -   &	-   & 65.30 & 73.50	&   -   &   -	& 76.80 \\    
        MAUM\cite{liu2022learningMAUM}          & CVPR 22  & 71.70 &   -   &	-   & 68.80 & 77.0	&   -   &   -	& 81.9 \\
        DSCNet \cite{r51zhang2022dual}          & TIFS 22  & 73.89 & 96.27 & 98.84 & 69.47 & 79.35 & 98.32 & 99.77 & 82.65 \\
        MSCLNet\cite{zhang2022modalityMSCL}     & ECCV 22  & 76.99 & 97.93 & 99.18 & 71.64 & 78.49 & 99.32 & 99.91 & 81.17\\
         \midrule
        GUR  \cite{yang2023towardsGUR}          & ICCV 23  & 63.51 &   -   &	-   & 61.63 & 71.11	&   -   &   -	& 76.23 \\        
        PMT  \cite{rPMTlu2023learning}          & AAAI 23  & 67.53 & 95.36 & 98.64 & 51.86 & 71.66 & 96.73 & 99.25 & 76.52 \\
        DEEN  \cite{r47zhang2023diverse}        & CVPR 23  & 74.70 & 97.60 & 99.20 & 71.80 & 80.30 & \textbf{99.00} & 99.80 & 83.30 \\      
        MUN  \cite{yu2023modalityMUN}           & ICCV 23  & 76.24 & 97.84 &   -   & 73.81 & 79.42 &   -   & 98.09 & 82.06 \\  
        SGIEL \cite{r48feng2023shape}           & CVPR 23  & 77.12 & 97.03 & 99.08 & 72.33 & 82.07 & 97.42 & 98.87 &	82.95 \\
        PartMix \cite{r52kim2023partmix}        & CVPR 23  & 77.78 &   -   &	-   & \textbf{74.62} & 81.52	&   -   &   -	& 84.83 \\ 
        \midrule
        \textbf{MSCMNet}           &   \textbf{Ours}  & \textbf{78.53} & \textbf{97.51} &  \textbf{99.23} & 74.20 & \textbf{83.00}	& 98.99 & \textbf{99.80} & \textbf{85.54} \\ 
      
        \bottomrule

    \end{tabular}

    }

\end{table*}
\begin{table}\footnotesize
    \centering
    \caption{COMPARISON WITH THE STATE-OF-THE-ARTS ON REGDB DATASET. RANK-R ACCURACY (\%) AND MAP(\%)}
    \setlength{\tabcolsep}{1.8mm}{
    \begin{tabular}{@{}l|c|cc|cc@{}}   
        \toprule
        \multirow{2}*{Method} & \multirow{2}*{Publish} & \multicolumn{2}{c}{V\ \ -- \ I} & \multicolumn{2}{c}{I\ --\ V} \\
          &       & Rank-1   & mAP & Rank-1   & mAP \\          
        \midrule
        Zero-Pad \cite{r26wu2017rgb}            & ICCV 17  & 17.8 & 18.9 & 16.7 & 17.9 \\
        HCML  \cite{rHCMLye2018hierarchical}    & AAAI 18  & 24.4 & 20.0 & 21.7 & 22.2 \\
        AliGAN \cite{r45AliGANwang2019rgb}      & ICCV 19  & 57.9 & 53.6 & 56.3 & 53.4 \\
        DDAG  \cite{r41ye2020dynamicDDAG}       & ECCV 20  & 69.3 & 63.4 & 68.0 & 61.8 \\ 
        AGW  \cite{rAGWye2021deep}              & TPAMI 21 & 70.0 & 66.3 & 70.4 & 65.9 \\
        LbA  \cite{r35park2021learningLbA}      & ICCV 21  & 74.1 & 67.6 & 72.4 & 65.4 \\
        CAJ  \cite{r38ye2021channel}            & CVPR 21  & 85.0 & 79.1 & 84.7 & 77.8  \\
         \midrule
        DML    \cite{RDMLzhang2022dual}         & TCSVT 22 & 77.6 & 84.3 & 77.0 & 83.6\\
        SPOT \cite{rsportchen2022structure}     & TIP 22   & 80.3 & 72.4 & 79.3 & 72.2 \\
        FMCNet  \cite{r39zhang2022fmcnet}       & CVPR 22  & 89.1 & 84.4 & 88.3 & 83.8 \\
        DCLNet  \cite{r34sun2022notDCL}         & ACM MM 22& 81.2 & 74.3 & 78.0 & 70.6 \\    
        MAUM\cite{liu2022learningMAUM}          & CVPR 22  & 87.9 & 85.1 & 87.0 & 84.3
 \\
        DSCNet \cite{r51zhang2022dual}          & TIFS 22  & 85.4 & 77.3 & 83.5 & 75.2 \\
        MSCLNet\cite{zhang2022modalityMSCL}     & ECCV 22  & 84.1 & 80.1 & 83.8 & 78.3 \\
         \midrule
        GUR  \cite{yang2023towardsGUR}          & ICCV 23  & 75.0 & 69.9 & 73.9 & 70.2 \\        
        PMT  \cite{rPMTlu2023learning}          & AAAI 23  & 84.8 & 76.5 & 84.1 & 75.1 \\
        PartMix \cite{r52kim2023partmix}        & CVPR 23  & 85.6 & \textbf{82.2} & 84.9 & \textbf{82.5 } \\

        \midrule
        \textbf{MSCMNet}                      &  \textbf{Ours}  & \textbf{90.4}  &  81.2 & \textbf{87.7}	&  78.2 \\     
        \bottomrule
    \end{tabular}

    }

\end{table}

\subsection{Implementation Details}
Our model is implemented using the PyTorch library and trained on an NVIDIA 3090 GPU. Firstly, we use AGW\cite{rAGWye2021deep} as our baseline. Additionally, each image is resized into 384 × 192 with zero-padding, random horizontal flipping, and random erasing\cite{zhong2020random} for data augmentation. We utilize the channel exchange technique from CAJ\cite{r38ye2021channel} to eliminate the channel information of RGB and employ our proposed IR channel expansion technique to incorporate channel information into IR images. During the training phase, we randomly selected 4 VIS images and 4 IR images from 6 identities. We utilize the SGD optimizer with a momentum of 0.9 and a weight decay of $ 5\times 10^{-4}$. The learning rate is decreased by a factor of 10 at epochs 30, 90, and 120 during the training process. We train the model for 150 epochs.

\begin{table}\footnotesize
    \centering
    \caption{COMPARISON WITH THE STATE-OF-THE-ARTS ON LLCM DATASET. RANK-R ACCURACY (\%) AND MAP(\%)}
    \setlength{\tabcolsep}{1.8mm}{
    \begin{tabular}{@{}l|c|cc|cc@{}}   
        \toprule
        \multirow{2}*{Method} & \multirow{2}*{Publish} & \multicolumn{2}{c}{V\ \ -- \ I} & \multicolumn{2}{c}{I\ --\ V} \\
          &       & Rank-1   & mAP & Rank-1   & mAP \\          
        \midrule

        DDAG  \cite{r41ye2020dynamicDDAG}       & ECCV 20  & 48.0 & 52.3 & 40.3 & 48.4 \\    
        LbA  \cite{r35park2021learningLbA}      & ICCV 21  & 50.8 & 55.6 & 43.8 & 53.1\\
        AGW  \cite{rAGWye2021deep}              & TPAMI 21 & 51.5 & 55.3 & 43.6 & 51.8 \\
        CAJ  \cite{r38ye2021channel}            & CVPR 21  & 56.5 & 59.8 & 48.8 & 48.8  \\
        MMN   \cite{MMNzhang2021towards}        & ACM MM 21  & 59.9 & 62.7 & 52.5 & 58.9  \\
        DART  \cite{DARTyang2022learning}       & CVPR 22  & 60.4 & 63.2 & 52.2 & 59.8  \\ 
        DEEN  \cite{r47zhang2023diverse}        & CVPR 23  & 62.5 & 65.8 & 54.9 & \textbf{62.9} \\            

        \midrule
        \textbf{MSCMNet}                      &  \textbf{Ours}  & \textbf{63.9}  & \textbf{66.1} & \textbf{55.1}	& 60.8 \\      
        \bottomrule
    \end{tabular}

    }

\end{table}

\subsection{Comparison with the State-of-the-Art Methods}
From the experiments on the SYSU-MM01 dataset shown in Tab. \uppercase\expandafter{\romannumeral1}, it can be observed that the proposed MSCMNet outperforms other state-of-the-art methods, achieves  78.53\% Rank-1 and 74.20\% mAP in the All-search mode, and 83.00\% Rank-1 and 85.54\% mAP in the Indoor-search mode. MSCMNet extracts more discriminative features by exploring the semantic correlations between features at different scales. In most VI Re-ID works, such as DSCNet \cite{r51zhang2022dual} and SGIEL \cite{r48feng2023shape}, the focus is on utilizing shape-erased or channel information to mine consistency between semantics. In contrast, MSCMNet better exploits the correlations of intra-modal and inter-modal semantic information achieved at multiple scales. Compared to methods based on data augmentation or auxiliary mode to reduce modality discrepancies, such as CAJ\cite{r38ye2021channel} and AliGAN\cite{r45AliGANwang2019rgb}, MSCMNet takes into account the semantic inconsistency caused by different data augmentation. Therefore, MSCMNet is more effective. From Tab. \uppercase\expandafter{\romannumeral2}, it is evident that MSCMNet also outperforms recent works on the RegDB dataset, with 90.4\% Rank-1 and 81.2\% mAP from visible to infrared mode as well as 87.7\% Rank-1 and 78.2\% mAP from infrared to visible mode. This reveals that achieving semantic consistency at the same scale is more challenging, and utilizing multi-scale semantic correlations for extracting discriminative features is a more effective approach. Finally, Tab. \uppercase\expandafter{\romannumeral3} demonstrates that MSCMNet also performs exceptionally well on the LLCM dataset.
\begin{table}\footnotesize
    \centering
    \caption{ABLATION STUDY OF QFE, MIMB, QCT ON THE all-search MODE OF SYSU-MM01 DATASET. RANK-R ACCURACY(\%) AND MAP(\%) ARE
REPORTED.}
    \setlength{\tabcolsep}{1.8mm}{
    \begin{tabular}{@{}c|c|c|c|c|c|c|c|c@{}}   \hline
        \multicolumn{5}{c|}{Method} & \multicolumn{4}{c}{SYSU-MM01} \\\hline
         BASE   &   QFE    & MIMB   & $\mathcal{L}_{tri}$ & $\mathcal{L}_{qct}$ &r-1&r-10&r-20&mAP \\\hline        
        \Checkmark    &             &          &\Checkmark&          &69.9 &95.7 &98.5 &66.9      \\\hline
        \Checkmark    & \Checkmark  &          &\Checkmark&          &73.0 &96.3 &98.8 &69.3      \\\hline
        \Checkmark    &             &\Checkmark&\Checkmark&          &74.3 &95.7 &98.4 &69.8      \\\hline
        \Checkmark    & \Checkmark  &\Checkmark&\Checkmark&          &75.1 &97.2 &98.9 &70.3      \\\hline
        \Checkmark    & \Checkmark  &          &          &\Checkmark&76.5 &97.2 &99.0 &71.8      \\\hline
        \Checkmark    &             &\Checkmark&          &\Checkmark&77.1 &97.4 &99.1 &71.8      \\\hline
        \Checkmark    & \Checkmark  &\Checkmark&          &\Checkmark&\textbf{78.5} &\textbf{97.5} &\textbf{99.2} &\textbf{74.2}     \\\Xhline{1pt}
    \end{tabular}

    }
\end{table}

\begin{figure}
\centering
\includegraphics[width=3.5in]{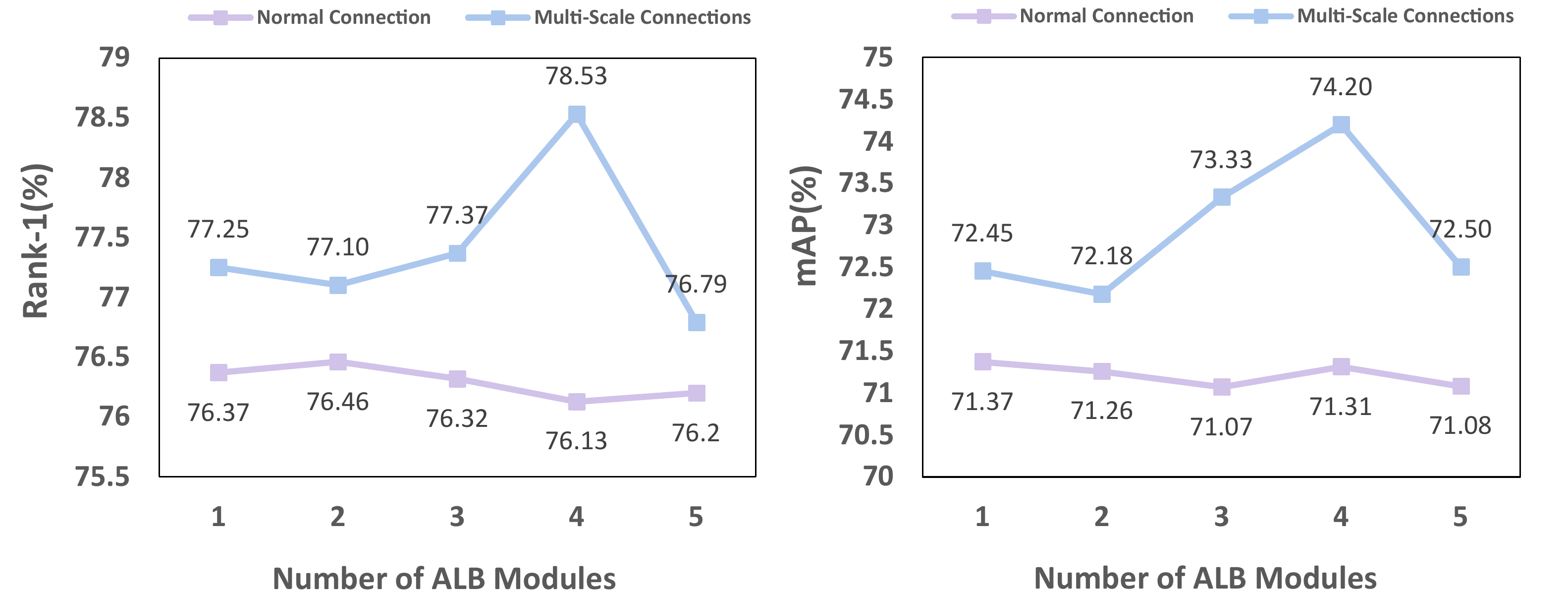}
\caption{Ablation study for MIMB with different numbers of the ALB and the effectiveness of multi-scale structures.}
\label{Fig_5}
\end{figure}
\subsection{Ablation Study and Analysis}
In this section, we conducted an ablation study to evaluate the contribution of each component in our proposed MSCMNet. All experiments were performed on SYSU-MM01 under the all-search mode using the same baseline. We tuned the hyperparameters of each experiment carefully. The results are described as follows:
\subsubsection{Effectiveness of Each Component}
 Here, we analyze the key components of MSCMNet, including the quadruple-steam feature extractor (QFE), Multi-scale Information Correlation Mining Block (MIMB), and quadruple center triple loss (QCT). For a fair comparison, we utilize CAJ as our baseline (Base) for all experiments. As summarized in Tab. \uppercase\expandafter{\romannumeral4}, each component helps to boost performance. Starting from the baseline, adding QFE improves the performance, which indicates that QFE effectively extracts features with different semantic information. This improvement can be attributed to the introduction of diverse information through data augmentation. When the MIMB is added, We observed considerable improvements in performance, confirming that MIMB effectively mines the semantic correlations between cross-modal features. It addresses the challenge of inconsistent semantic information at the same scale. Thereby enabling the extraction of comprehensive from diverse semantic dimensions. Finally, experiments indicate that QCT extracts more discriminative features of the person and enables the extracted features to incorporate the most stable information from different data augmentation. The combination of these components has resulted in performance gains for MSCMNet.

\begin{figure}[t]
\centering
\includegraphics[width=2.3in]{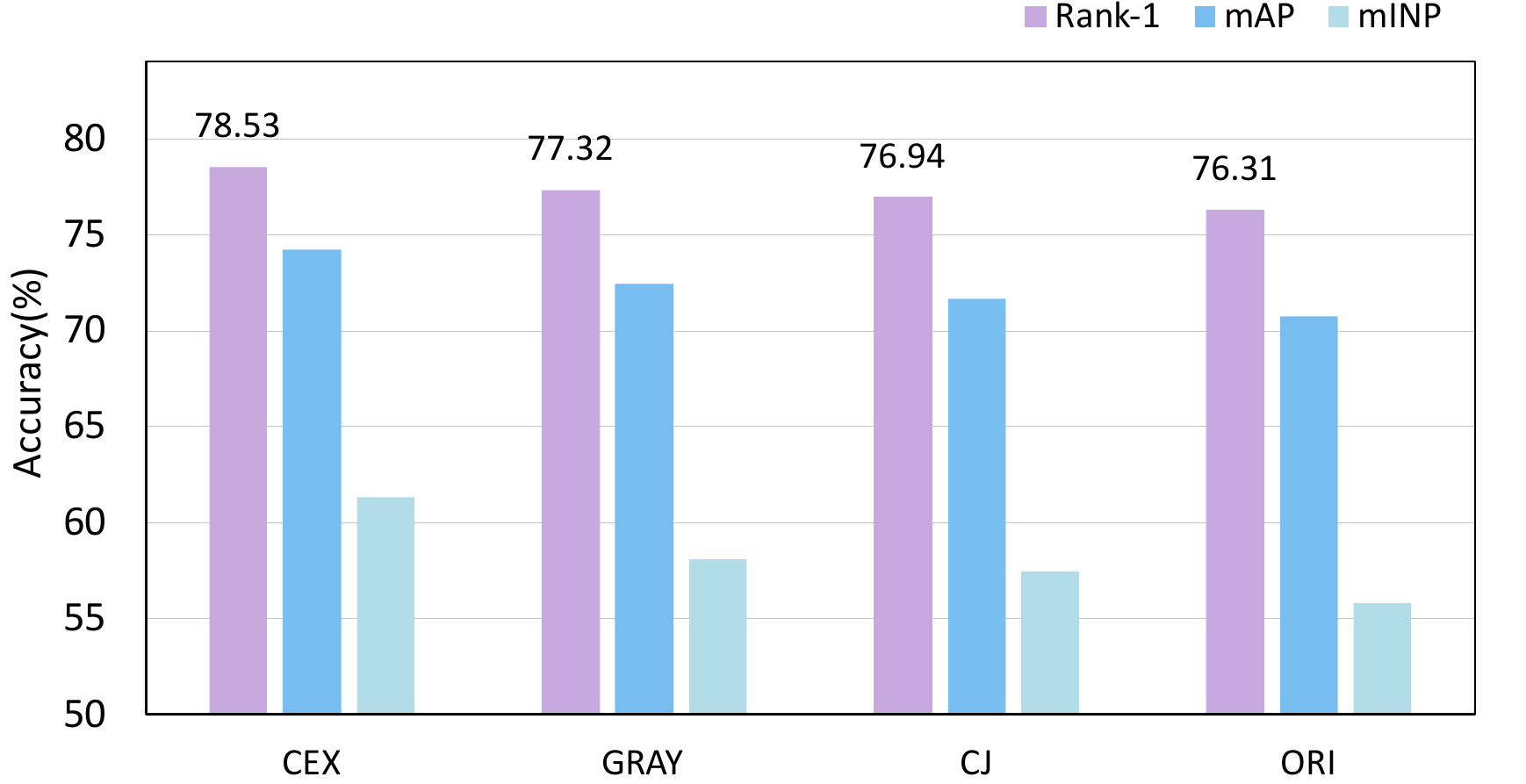}
\caption{Performance analysis of the extra branch for visible modality with different data augmentation.}
\label{Fig_6}
\end{figure}

\begin{figure}[t]
\centering
\includegraphics[width=3.5in]{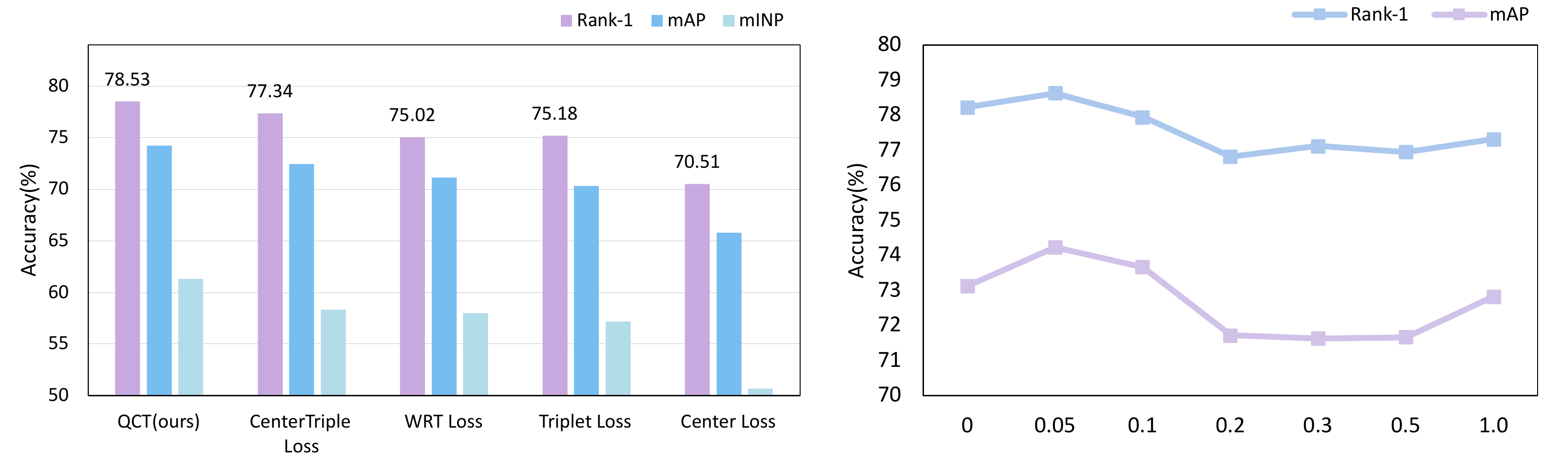}
\caption{Performance analysis of different loss functions on the left and effects of trade-off parameters $\alpha$ on Quadruple Center Triple Loss on the right.}
\label{Fig_7}
\end{figure}

\subsubsection{Effectiveness of multi-scale structure MIMB and the number of ALB layers}

Firstly, the proposed MIMB is composed of ALB. As shown in Fig. 5, the results show that the best performance is obtained using 4-layer ALB. This is because 4-layer ALB can encompass all scale information of the Network shallow layer. On the one hand, using 5-layer ALB would result in information redundancy of shallow layer features, which is not conducive to network training. On the other hand, employing 3 or fewer layers ALB leads to the loss of semantic information on a specific scale, causing the MIMB to lose the ability to explore the semantic correlations in that part of the feature. Furthermore, to validate the effectiveness of the multi-scale structure, we conducted experiments on MIMB with the multi-scale structure removed. The results demonstrate that regardless of the number of layers in ALB, the performance is superior when the multi-scale structure is integrated. Compared to existing pyramid-based multi-scale approaches\cite{zhang2021multiMMM1,liu2022learnMMM2,wen2022crossMM3}, our work designs an innovative multi-scale structure that enables a more comprehensive feature exploration while considering a richer set of modality information. As a result, we achieve higher accuracy. Moreover, simply stacking ALB modules without the multi-scale structure leads to poor performance and results in the network failing to fit adequately. This observation further confirms the effectiveness of our MIMB.

\subsubsection{The impact of different data augmentation on the additional RGB branch}
As shown in Fig. 6, we conducted tests on data augmentation for the extra visible branch, and it was observed that the channel exchange (CEX)\cite{r38ye2021channel} technique achieved the best results. This is because channel exchange directly selects channel information as the network input without any processing and keeps the original channel information of the image. Although global information contains channel information, it actually emphasizes different semantic dimensions. This is also because our model aims to explore the implicit semantic correlations across diverse information that is present in the data. Based on experimental results, we ultimately selected channel exchange as the method to eliminate channel information from RGB images.

\begin{figure*}[!t]
\centering
\includegraphics[width=7in]{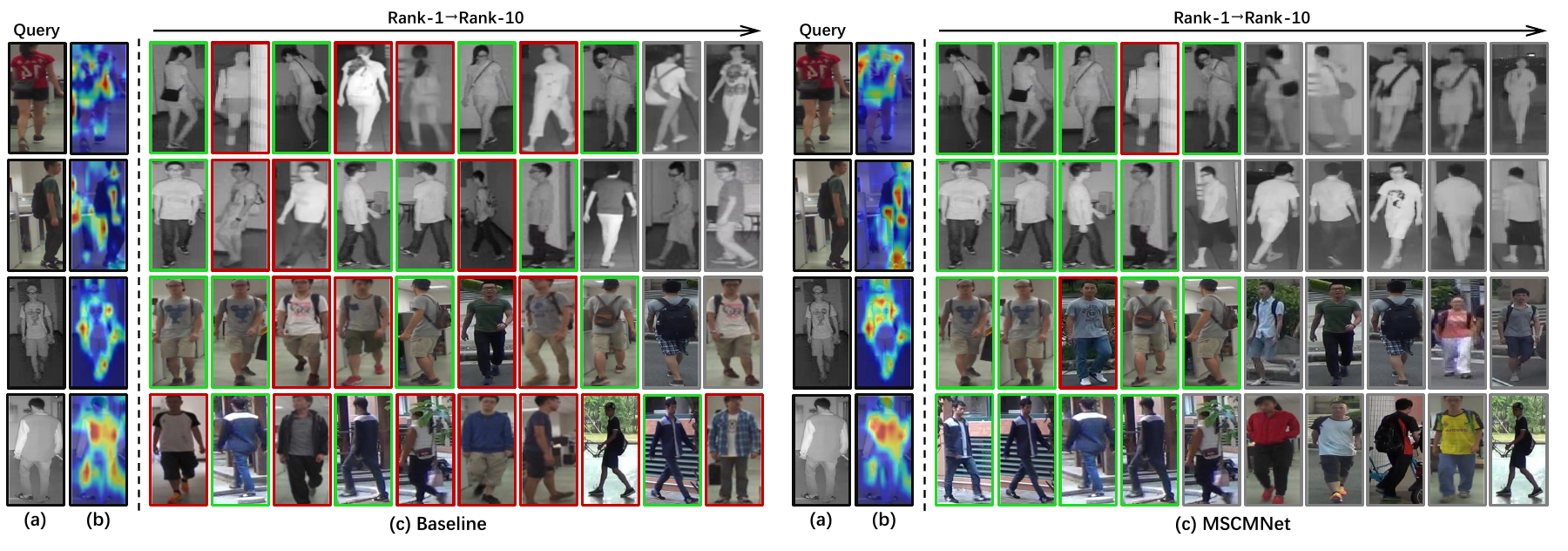}%
\hfil
\caption{Illustration of person retrieval results and Heat maps. (a) Query images. (b) Heat map. (c) Top 10 retrieval results. The left side is the retrieval result of our MSCMNet, and the right side is the baseline retrieval result.}
\label{Fig_8}
\end{figure*}

\begin{figure}
\centering
\includegraphics[width=3.4in]{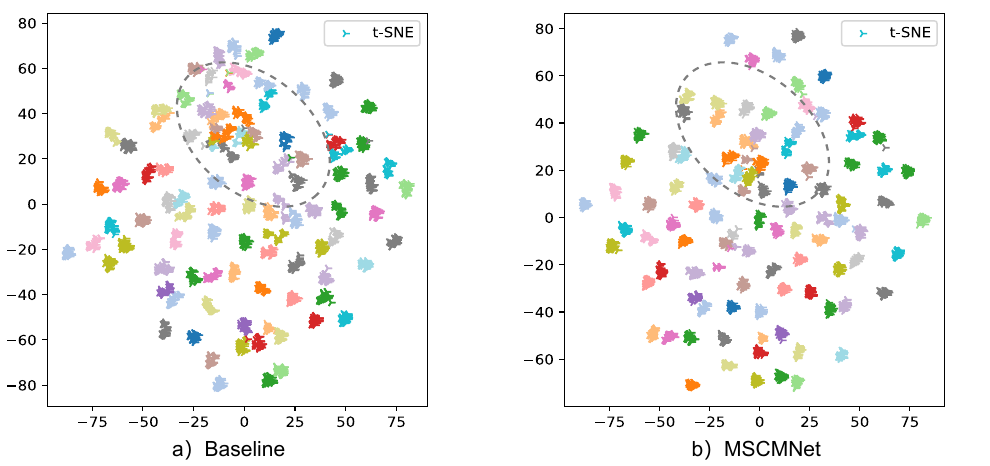}
\caption{T-SNE visualization result of baseline with MSCMNet.}
\label{Fig_9}
\end{figure}

\begin{figure}
\centering
\includegraphics[width=3.4in]{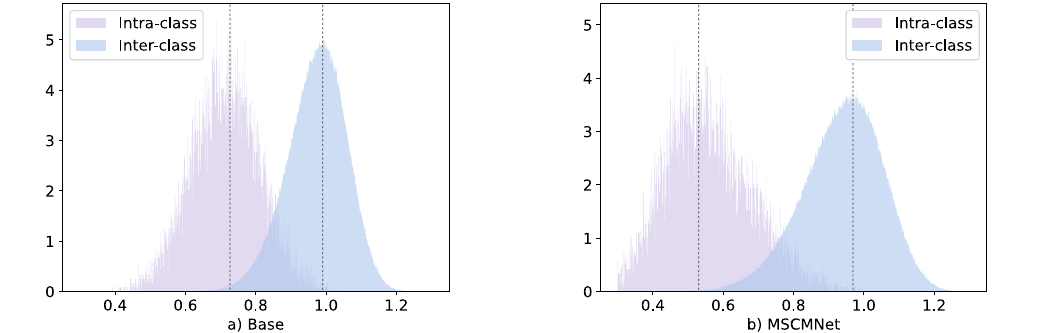}
\caption{The feature distances of intra-and-inter classes visualization.}
\label{Fig_10}
\end{figure}

\subsubsection{Compare the effects of different loss functions}
The objective of designing the Quadruple Center Triple Loss (QCT) is to enable the network to effectively leverage information from different dimensions, which enables improved constraint effects. To demonstrate the effectiveness of the loss, we conducted performance testing by comparing it with popular loss functions. As shown in Fig. 7 left, we test the Center Loss, Triple Loss, WRT, and Center Triple Loss on the MSCMNet, and the results demonstrate that our QCT achieves greater improvements. This can be attributed to the fact that the features extracted from different data emphasize different dimensions of semantics. By utilizing QCT, we can effectively explore these features to take advantage of our network. Fig. 7 right demonstrates the impact of parameter $\alpha$ in QCT and we set the value of parameter $\alpha$ to $0.05$.

\subsection{Visualization Analysis}
To analyze the visual effectiveness of our proposed model, we present several representative visual examples. As shown in Fig. 8(b), we employ Grad-CAM to generate attention maps for the query images. Compared to the heatmaps extracted by the baseline model displayed on the left, the heatmaps generated by MSCMNet on the right exhibit a stronger focus on identity-related features and contain more focus points within the regions of individuals. This observation suggests that our design of MSCMNet captures a broader range of person-specific characteristics. Additionally, Fig. 8(c) provides the top-10 retrieval results. It proves that MSCMNet learns more comprehensive identity discrimination between different classes compared with the baseline. As shown in Fig. 9, the learned identity discrimination of the baseline model is visualized using t-SNE, revealing that feature embeddings of the same identity are dispersed and difficult to distinguish. For MSCMNet, since we design QCT to achieve a stronger constraint effect, each identity can be cast to a more compact and distinguishable distribution. Finally, We visualize the intra-and-inter identity feature distance in Fig. 10. In the all-search modes, mean values for the feature distances of intra-class decline prove that MSCMNet successfully reduces the modality divergence compared with the baseline.

\section{Conclusion}
In this paper, we propose a novel VI-ReID framework called MSCMNet. It focuses on exploring the correlations between different modality semantic features at multiple scales, ensuring that the extracted features encompass comprehensive modality and identity discriminative information. We improve the issue of existing networks that lose valuable information during the feature extraction process. Our proposed QFE and MIMB techniques effectively extract and utilize features from multiple semantic dimensions, leading to improved accuracy and robustness of the model. It is worth noting that our MIMB structure can be further combined with other existing VI-ReID models for further investigation. Extensive experimental results validate the outstanding performance of MSCMNet and the effectiveness of its components. In the future, we are driven to better explore the implicit correlations among semantic information of multiple scales in other cross-modal tasks.
\bibliographystyle{unsrt} 
\bibliography{MSCMNet}
\end{document}